\documentclass[letterpaper, 10 pt, conference]{ieeeconf}  

\IEEEoverridecommandlockouts                              

\usepackage{multicol}
\usepackage[bookmarks=true]{hyperref}
\usepackage{amsfonts,bm,nccmath}
\usepackage{graphicx}
\usepackage{subcaption}

\title{\LARGE \bf
Generative grasp synthesis from demonstration using parametric mixtures
}

\author{Ermano Arruda$^{1}$, Claudio Zito$^{1}$, Mohan Sridharan$^{1}$, Marek Kopicki$^{1}$ and Jeremy L. Wyatt$^{1}$
\thanks{\footnotesize $^{1}$School of Computer Science, University of Birmingham, B15 2TT, UK  \tt (exa371,c.zito,M.Sridharan,msk,jlw)@cs.bham.ac.uk}%
\thanks{\footnotesize Accepted for publication in the RSS workshop on Task-Informed Grasping (TIG-II), 2019.}%
}

\begin{document}

\maketitle

\begin{abstract}

We present a parametric formulation for learning generative models for grasp synthesis from a demonstration. We cast new light on this family of approaches, proposing a parametric formulation for grasp synthesis that is computationally faster compared to related work and indicates better grasp success rate performance in simulated experiments, showing a gain of at least 10\% success rate $(p < 0.05)$ in all the tested conditions. The proposed implementation is also able to incorporate arbitrary constraints for grasp ranking that may include task-specific constraints. Results are reported followed by a brief discussion on the merits of the proposed methods noted so far.


\end{abstract}


\section{Introduction}

We address the problem of modelling generative models for grasp synthesis using parametric representations for probability densities. In recent decades, considerable effort has been put towards tackling the problem of grasping \cite{shimoga1996robot,bicchi2000a,curtis2008a,ben-amor2012a,rietzler2013a,detry2013c,hjelm2014a,saxena2008b,bohg2011b,kopicki2015,Pas2015,Gualtieri2016,Levine2016}.



Generative approaches for grasp synthesis have demonstrated to be more versatile since they can be used to generate complete hand configurations in novel contexts, e.g. \cite{Montesano2008,Song2011,Morrison2018}.
Kopicki et al \cite{kopicki2015} have proposed a non-parametric technique to learn generative models from demonstration for dexterous grasping via Kernel density estimation (KDE). Such approach makes use of KDEs to approximate probability densities over the special Euclidian group - $SE(3)$, as well as hand configuration models encoding the shape of the hand for a given demonstration. The original approach has been shown to be able to generalise demonstrated grasps to novel query object point clouds, without requiring prior knowledge about object pose, shape (such as CAD mesh models) or dynamic properties such as friction coefficients, mass distribution, among others.

Although KDEs are very appealing, one drawback of learning such non-parametric models is the computational time when evaluating the likelihood of samples. If a KDE has $K$ kernels and one wishes to evaluate the likelihood of $N$ data points under the model, the time complexity for computing the data likelihood is $O(K N)$. Noting that for KDEs $K$ is typically large, in the order of hundreds, since commonly a kernel is placed on every training data point. This large number of kernels makes the final time complexity dependent on the size of the training data set used to approximate a desired density. In contrast, parametric mixtures usually need a smaller number for $K$ to effectively approximate a probability density. Once the model parameters are learnt via Expectation Maximisation (EM) \cite{Bishop2007PatternRA}, the runtime for evaluating the likelihood of query data points is not dependent on the size of the data set used for training, but only on the fixed number chosen for $K$. Thus, although the complexity for data likelihood computation is the same in principle, in practice parametric mixtures are orders of magnitude faster than KDEs since $K$ is rarely greater than 10 for most scenarios. 

\begin{figure}[!t]
\centering
\includegraphics[trim={0cm 3cm 8cm 0cm}, clip,width=0.6\columnwidth]{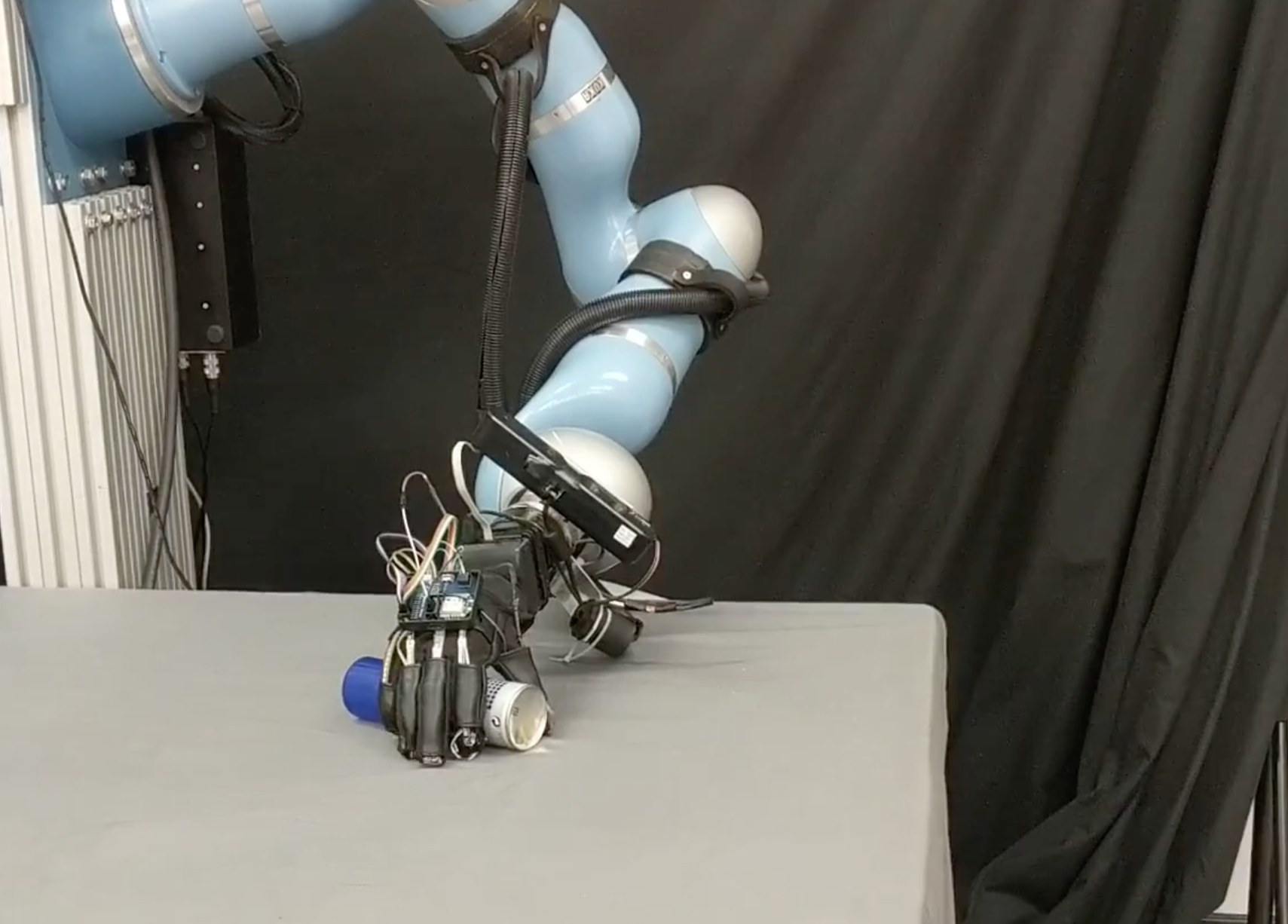}
\caption{{\small Preliminary deployment of KDE and GMM-based grasp synthesis methods on the Boris robot platform.}}
\label{fig:cylinder_grasp} 
\vspace{-18pt}
\end{figure}

This work re-interprets the approach in \cite{kopicki2015} for grasp synthesis using a parametric mixture formulation for density modelling. We show the benefits for grasp generation time and performance in simulation under different sensor noise conditions, also showing deployment on a robot platform.

%


First, we proceed to describe the basic representations utilised throughout this paper for modelling probability densities, followed by the description of our approach. Finally, we present our experimental results in simulation and show preliminary deployment of the approach on a robot platform.

\section{Representations}

\begin{figure}[!t]
\centering
\includegraphics[width=0.45\columnwidth]{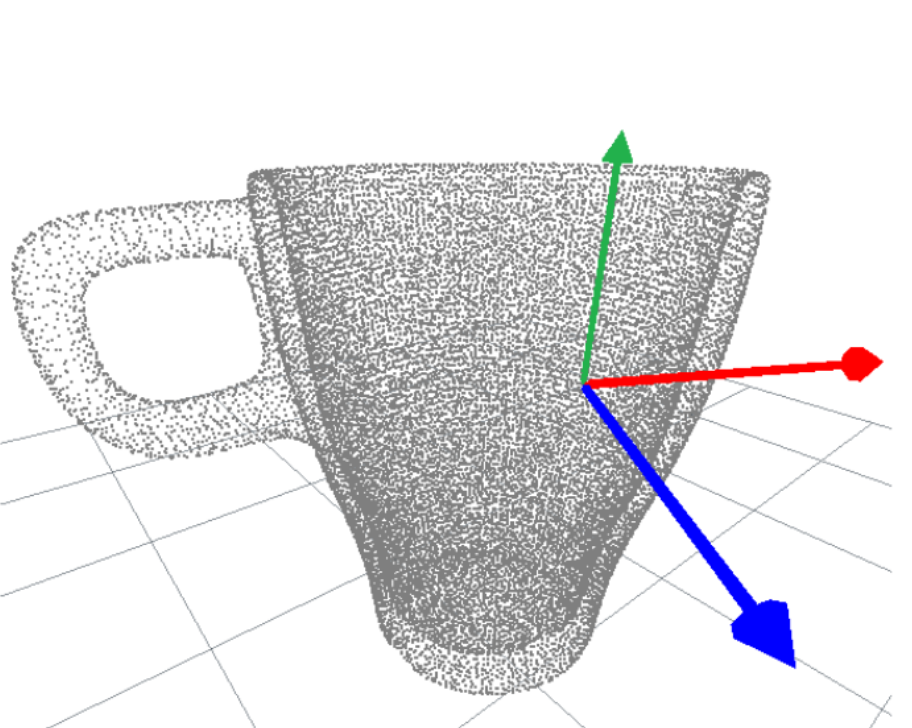}
\caption{An example point cloud of a mug and feature representation. For a given point $\mathbf{p}$, the axis in blue is the surface normal pointing outwards the object, the direction of the first principal curvature $k_1$ is depicted as the horizontal red axis, and the direction of the second principal curvature $k_2$ is depicted as the green vertical axis. This right-handed frame is constructed by taking the cross-product between the normal and the first principal curvature direction. Together with the point $\mathbf{p}$, these three axis define a rigid body pose, whose position and quaternion representation are given by $\mathbf{v}=(\mathbf{p},\mathbf{q})$. Therefore, each point of the object has a rigid frame attached in similar fashion, although in the picture we chose to highlight only one to avoid clutter.}
\label{fig:point_cloud_and_feat}
\end{figure}

We are interested in modelling joint probability densities over rigid body transformations belonging to the $SE(3)$ and arbitrary feature vectors in $\mathcal{R}^n$. Thus, realisations of the random variable we intend to model lie in the space $\mathbf{x} \in SE(3) \times \mathcal{R}^n$. Concretely, the special Euclidean group is the cartesian product of members of the special orthogonal group $SO(3)$ (the group of rotations) and members of $\mathcal{R}^3$ representing translations, thus $SE(3) = \mathcal{R}^3 \times SO(3)$. If we choose to represent rotations with unit quaternions, then $\mathbf{x}=(\mathbf{p}, \mathbf{q}, \mathbf{r})$ is a $d=7 + n$ dimensional vector, where $\mathbf{p} \in \mathcal{R}^3$, $\mathbf{q} \in \mathcal{R}^4$ and $\mathbf{r} \in \mathcal{R}^n$. In this work, the same type of feature used by \cite{kopicki2015} will be utilised, which are the eigen-values $k_1$ and $k_2$ of the principal curvatures for a given point $\mathbf{p}$ from an object point cloud. In Fig. \ref{fig:point_cloud_and_feat} this type of feature is illustrated. Thus, concretely, the feature vector is represented as $\mathbf{r}=[k_1,k_2] \in \mathcal{R}^2$.

We will see that we are able to learn probability densities over this manifold using data acquired from demonstration where a data set $\mathcal{D}$ is acquired, such that

\begin{ceqn}
\begin{equation}
\label{eq:data}
\mathcal{D} = \{ \mathbf{x_j} | \mathbf{x_j} \in \mathcal{R}^3 \times \mathcal{SO}(3) \times \mathcal{R}^n \}_{j=1}^{J_D}
\end{equation}
\end{ceqn}
\noindent where $\mathbf{x}_j \sim \mathbf{pdf}(\mathbf{p}, \mathbf{q}, \mathbf{r})$.

In particular, we will represent quaternions noting their relationship with angular velocities $\bm{w} \in \mathcal{R}^3$ through the logaritimic and exponential maps defined by Eq. \ref{eq:log_map} and \ref{eq:exp_map}, as similarly noted by \cite{Ude2014}. Given a unit quaternion $\mathbf{q} = [\mathbf{q}_v, q_w]^T$, we define its logarithmic map as:

\begin{ceqn}
\begin{equation}
\label{eq:log_map}
\bm{\omega} = \log(\mathbf{q}) = \left\{
        \begin{array}{ll}
            \arccos(q_w) \frac{\mathbf{q}_v}{ \| \mathbf{q}_v \| } & \quad \mathbf{q}_v \neq \mathbf{0} \\
            {[0,0,0]^T} & \quad otherwise
        \end{array}
    \right.
\end{equation}
\end{ceqn}
\\
And conversely, the exponential map is given by:

\begin{ceqn}
\begin{equation}
\label{eq:exp_map}
\mathbf{q} = \exp(\bm{\omega}) = \left\{
        \begin{array}{ll}
            {[\sin(\| \bm{\omega}\| ) \bm{\omega}, \cos(\| \bm{\omega}\| )]^T} & \quad \bm{\omega} \neq \mathbf{0} \\
            {[0,0,0,1]^T} & \quad otherwise
        \end{array}
    \right.
\end{equation}
\end{ceqn}

Using these mappings, we will fit parametric mixtures to model joint distributions over $\mathbf{x} = (\mathbf{p}, \bm{\omega}, \mathbf{r})$, where the quaternion representation for $\bm{\omega}$ is readily given by the exponential map $\mathbf{q} = \exp(\bm{\omega})$.

\subsection{Probability density approximation using parametric mixtures}

The density $\mathbf{pdf}(\mathbf{x})$ is approximated using a parametric Gaussian mixture model (GMM) modified so as to internally take care of the appropriate exponential and logarithmic mappings for quaternions. It is then defined as follows:

\begin{ceqn}
\begin{equation}
\label{eq:density_mixture}
\mathbf{pdf}(\mathbf{x}) = \sum_{j=1}^{K} w^j \mathcal{N}(\mathbf{x} | \bm{\mu_x}^j, \bm{\Sigma_x}^j)
\end{equation}
\end{ceqn}

For convenience, let $\mathbf{u}=(\mathbf{p},\mathbf{q})$, i.e. a rigid body transformation. Thus, with ${\bm{\Sigma_x}^j}^{-1} = \bm{\Lambda_x}^j$, we denote:

\begin{ceqn}
\begin{subequations}
\begin{align}
\bm{\mu_x}^j &= (\bm{\mu_u}^j, \bm{\mu}_r^j), \label{eq:mu_var} \\
\bm{\Sigma_x} &= \begin{pmatrix}
  \begin{matrix}
  \bm{\Sigma_{uu}}^j & \bm{\Sigma_{ur}}^j  \\
  \bm{\Sigma_{ur}}^j & \bm{\Sigma_{rr}}^j
  \end{matrix}
\end{pmatrix} , \bm{\Lambda_x}^j = \begin{pmatrix}
  \begin{matrix}
  \bm{\Lambda_{uu}}^j & \bm{\Lambda_{ur}}^j  \\
  \bm{\Lambda_{ur}}^j & \bm{\Lambda_{rr}}^j
  \end{matrix}
\end{pmatrix} 
\end{align}
\end{subequations}
\end{ceqn}




In this work the density $p( \mathbf{u} | \mathbf{r})$ is a conditional probability density of rigid body transformations given a contact feature $\mathbf{r}$, and is also modelled as a Gaussian mixture model. The parameters of this conditional mixture are obtained in closed form from Eq. \ref{eq:density_mixture} as (see \cite{Bishop2007PatternRA} for an overview):

\begin{ceqn}
\begin{subequations}
\begin{align}
\bm{\mu_{u|r}}^j &= \bm{\mu_u}^j - \bm{\Sigma_{uu}}^j \bm{\Lambda_{ur}}^j (\mathbf{r} - \bm{\mu_r}^j),  \label{eq:mu_var_ur} \\
\quad \nonumber \\
\bm{\Sigma_{u|r}}^j &= \bm{\Sigma_{uu}}^j \label{eq:sigma_var_ur} \\
\quad \nonumber \\
\qquad p^j(\mathbf{u} | \mathbf{r}) &= \mathcal{N}(\mathbf{x} | \bm{\mu_{u|r}}^j , \bm{\Sigma_{u|r}^j}) \\
\quad \nonumber \\
\pi^j &= \frac{w^j \mathcal{N}(\mathbf{r} | \bm{\mu_r}^j, \bm{\Sigma_{rr}}^j)}{\sum_{j=1}^{K} \mathcal{N}(\mathbf{r} | \bm{\mu_r}^j, \bm{\Sigma_{rr}}^j)}
\end{align}
\end{subequations}
\end{ceqn}

Therefore, $p( \mathbf{u} | \mathbf{r})$ is expressed as  
\begin{ceqn}
\begin{equation}
\label{eq:density_mixture_rigid_body}
p( \mathbf{u} | \mathbf{r}) = \sum_{j=1}^{K} \pi^j p^j( \mathbf{u} | \mathbf{r})
\end{equation}
\end{ceqn}

The final density over features $\mathbf{r}$, i.e. $p(\mathbf{r})$, is also modelled as a Gaussian mixture and is obtained via marginalisation of Eq. \ref{eq:density_mixture} with respect to $\mathbf{u}$ such that:

\begin{ceqn}
\begin{equation}
\label{eq:density_mixture_features_marg}
p^j( \mathbf{r}) = \int \mathcal{N}(\mathbf{x} | \bm{\mu_x}^j, \bm{\Sigma_x}^j) du = \mathcal{N}(\mathbf{r} | \bm{\mu_r}^j, \bm{\Sigma_r}^j)
\end{equation}
\end{ceqn}

Thus, it follows that

\begin{ceqn}
\begin{equation}
\label{eq:density_mixture_features}
p(\mathbf{r}) = \sum_{j=1}^{K} w^j p^j( \mathbf{r}).
\end{equation}
\end{ceqn}

The mixture in Eq. \ref{eq:density_mixture} is learned using the expectation maximisation algorithm (EM) over contact data acquired from a grasp demonstration $\mathcal{D}= \{ \mathbf{x_j} | \mathbf{x_j} \in \mathcal{R}^3 \times \mathcal{SO}(3) \times \mathcal{R}^n \}$. The mixture formulation is more compact and less memory hungry. Once the density is learnt, we only need to keep its learnt parameters in memory and compute in closed form the additional probability densities in Eq. \ref{eq:density_mixture_features} and Eq. \ref{eq:density_mixture_rigid_body}. In addition, it needs fewer kernels to approximate the densities, therefore $K$ is typically smaller than 10. In contrast to the KDE approach, each data point becomes a the center of a kernel, which is of the order of hundreds per contact model learnt.

\section{Grasp synthesis using parametric contact models}

Having defined the basic representations for learning distributions using parametric mixtures, we proceed to describe the main components of what we refer to as the parametric grasp synthesis approach.

We will be learning joint probability densities over $\mathbf{x}=(\mathbf{p}, \mathbf{q}, \mathbf{r})$. For compactness we will refer to the position and orientation (pose) of a feature $\mathbf{r}$ as $\mathbf{v}=(\mathbf{p}, \mathbf{q})$, and therefore $\mathbf{x}$ can be written as

\begin{ceqn}
\begin{equation}
\label{eq:compact_x}
\mathbf{x} = (\mathbf{v},\mathbf{r}),
\end{equation}
\end{ceqn}

\noindent where $\mathbf{v} \in SE(3)$ and $\mathbf{r} \in \mathcal{R}^n$ is a feature vector.



\subsection{The object representation} At all times, the object intended to be grasped is represented by $O(\mathbf{v}, \mathbf{r})$. Concretely, it consists of the current object point cloud augmented with features $\mathbf{r} = [k_1, k_2]$ representing the principal curvatures at each point of the object. Finally, for a given point $\mathbf{p}$ from the point cloud, using corresponding eigenvectors of the principal curvatures and estimated normal at this given point, one can construct a frame $\mathbf{v} = (\mathbf{p},\mathbf{q})$. Hence, the object point cloud is then augmented to be a cloud of rigid body transformations associated with a features $\mathbf{r}$ as depicted by Fig. \ref{fig:point_cloud_and_feat}. 


\subsection{Contact model}

For a given hand link $L_i$, the conditional contact model density $M_i(\mathbf{u}|\mathbf{r})$ is modelled as a conditional probability density over finger link poses $\mathbf{u} \in SE(3)$ given contact surface features $\mathbf{r} \in \mathcal{R}^n$. This density is proportional to the likelihood of finding a finger link located at $\mathbf{u}$ with respect to the location $\mathbf{v}$ of a contact point with feature vector $\mathbf{r}$. Such model is constructed from the dataset $\mathcal{D}_O = \{\mathbf{x}_j\}_{j=1}^{K_O}$ as follows. Let $\mathbf{s}_i = (\mathbf{p}_i, \mathbf{q}_i) \in SE(3)$ be the pose of the link $L_i$ in world frame. The relative pose of link $L_i$ with respect to a feature pose $\mathbf{v}_j$ is given by

\begin{ceqn}
\begin{equation}
\label{eq:relative_pose_u_mixture}
\mathbf{u}_{ij} = (\mathbf{p_{ij}},\mathbf{q_{ij}}) = \mathbf{v}_j^{-1} \circ \mathbf{s}_i
\end{equation}
\end{ceqn}

Thus we are able to construct the contact model representing a probability distribution of relative poses with respect to contact features $\mathbf{r}_j$ as

\begin{ceqn}
\begin{equation}
\label{eq:contact_model_mixture}
M_i(\mathbf{u}|\mathbf{r}) \approx p( \mathbf{u} | \mathbf{r})
\end{equation}
\end{ceqn}

\noindent where $p( \mathbf{u} | \mathbf{r})$ is defined in Eq. \ref{eq:density_mixture_rigid_body} and is obtained via conditioning Eq. \ref{eq:density_mixture} learnt using the demonstration data set $\mathcal{D}_O$.

Furthermore, it is worth noting that $M_i$ is the contact model for link $L_i$ in proximity during demonstration to a set of features modelled by

\begin{ceqn}
\begin{equation}
\label{eq:contact_model_mixture_feature}
M_i(\mathbf{r}) \approx p( \mathbf{r})
\end{equation}
\end{ceqn}

\noindent where $M_i(\mathbf{r})$ is obtained via marginalisation of Eq. \ref{eq:density_mixture} as described by Eq. \ref{eq:density_mixture_features_marg}. The mixture in Eq. \ref{eq:density_mixture} is learnt with EM using the demonstration data set $\mathcal{D}_O = \{\mathbf{x}_j\}_{j=1}^{K_O}$, a data set consisting of data points $\mathbf{x}_j = (\mathbf{v}_j,\mathbf{r}_j)$ in proximity to finger link $L_i$.

Let $(\mathbf{\tilde{v}}, \mathbf{\tilde{r}}) \sim O(\mathbf{v}, \mathbf{r})$ be a sample from the object intended to be grasped represented by $O(\mathbf{v}, \mathbf{r})$. It follows that if were to compute the data likelihood of features $\mathbf{\tilde{r}}$ over $O(\mathbf{v}, \mathbf{r})$, then features with high data likelihood reflect higher affinity that this finger link $L_i$ could be positioned over $\mathbf{\tilde{r}}$ located at $\mathbf{\tilde{v}}$. The location of this finger link in world frame is given by $\mathbf{\tilde{s}} = \mathbf{\tilde{v}} \circ \mathbf{\tilde{u}}$, where $\mathbf{\tilde{u}} \sim M_i(\mathbf{u}|\mathbf{\tilde{r}})$, together with its weight representing this affinity $\tilde{w} = M_i(\mathbf{\tilde{r}})$, forming the n-uple $(\mathbf{\tilde{s}},\tilde{w})$.

\begin{figure}[!t]
\centering
\begin{minipage}{.5\columnwidth}
  \centering
  \includegraphics[trim={3cm 1cm 3cm 3cm},clip,width=.9\linewidth]{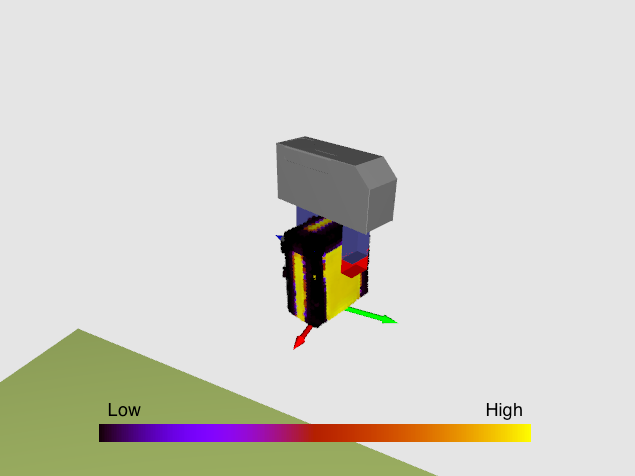}
\end{minipage}%
\begin{minipage}{.5\columnwidth}
  \centering
  \includegraphics[trim={3cm 1cm 3cm 3cm},clip,width=0.9\linewidth]{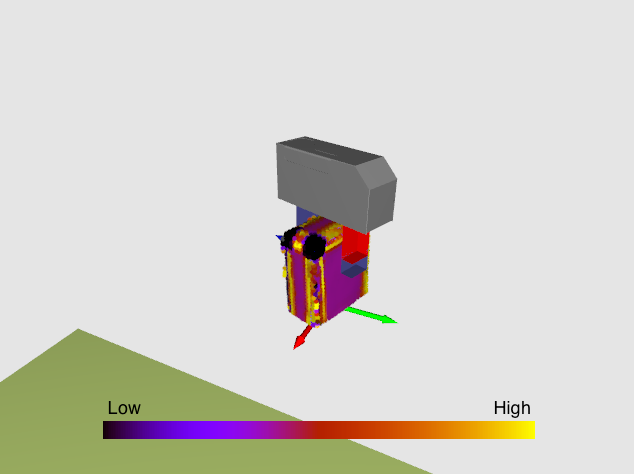}
%
\end{minipage}
\caption{\small{Left: contact model distribution for finger link $L_2$. This finger link has higher affinity to flat surfaces of the object. Right: contact model distribution for finger link $L_1$. This finger link has higher affinity to edge surfaces of the object.}}
\label{fig:cd_feat}
\vspace{-12pt}
\end{figure}

Depending on the probability distribution of features during demonstration, different finger links show different affinities, thus modelling different contact distributions. Figure \ref{fig:cd_feat} 
depicts this property for two different links of a WSG 50 Schunk gripper. For instance, in Fig. \ref{fig:cd_feat} (left), the link number $L_2$ is closer to flat patches of surface. By evaluating $M_2(\mathbf{\tilde{r}})$ over $(\mathbf{\tilde{v}}, \mathbf{\tilde{r}}) \sim O(\mathbf{v}, \mathbf{r})$, one is able to determine the regions of high and low data likelihood/affinity over the object surface represented by $O(\mathbf{v}, \mathbf{r})$, which in this case is a soup box. In Fig. \ref{fig:cd_feat} (right), the same process is repeated this time for $L_1$ . Note that $L_2$ prefers flat surfaces, whereas $L_1$ prefers corners, due to its closer proximity to features describing the edges of the object during demonstration. The features we chose to utilise are the same features described by $\cite{kopicki2015}$, which are the principal axis of curvature, computed using a PCA method. Thus, each feature $\mathbf{r} \in \mathcal{R}^2$ is a two dimensional vector containing the two largest eigen values approximating the curvatures of a given point of the object point cloud.

\subsection{Hand configuration model}

A probability density constructed from a demonstrated grasp trajectory containing starting and final joint configurations of the robot hand. Here, this model is constructed as originally proposed by \cite{kopicki2015}. Denoting $\mathbf{h}^t_e$ as the joint angles at the start of a given demonstrated grasp (pre-grasp configuration), and $\mathbf{h}^t_g$ the joint angles at contact with the object. Finally, let the set of hand configurations be $\mathcal{H}_c = \{ \mathbf{h}(\gamma) : \gamma \in [-\beta, \beta], \beta \in \mathcal{R}^+\}$, where $\mathbf{h}(\gamma) = (1 - \gamma) \mathbf{h}^t_g + \gamma \mathbf{h}^t_e$, the probability density $C(\mathbf{h}_c)$ is then non-parametrically approximated from this set via a KDE as

\begin{ceqn}
\begin{equation}
\label{eq:manipulator_model}
C(\mathbf{h}_c) = \sum_{\gamma \in [-\beta, \beta]} w(\mathbf{h}_c(\gamma)) \mathcal{N}_D(\mathbf{h}_c|\mathbf{h}_c(\gamma),\sigma_c) 
\end{equation}
\end{ceqn}

\noindent where $w(\mathbf{h}_c(\gamma)) = \exp(-\alpha \| \mathbf{h}_c(\gamma) - \mathbf{h}^t_g \|)$ and $\alpha \in \mathcal{R}^+$.

\subsection{Grasp synthesis for novel objects} 

\subsubsection{Contact query density}

From the learnt densities described above, given a query point cloud for a new object in its augmented representation $O(\mathbf{v}, \mathbf{r})$, a so called contact query density $Q_i(\mathbf{s})$ can be constructed for each link $L_i$. This probability density is constructed by first sampling $(\mathbf{\tilde{v}}, \mathbf{\tilde{r}}) \sim O(\mathbf{v}, \mathbf{r})$ from the object intended to be grasped. Next, $\tilde{\mathbf{u}}_i \sim M_i(\mathbf{u}|\mathbf{r})$ is sampled from our learnt contact model for finger link $L_i$ in Eq. \ref{eq:contact_model_mixture}, such that a set of $N_{Q_i}$ samples $\tilde{\mathbf{s}}_{ij} = \tilde{\mathbf{v}}_j \circ \tilde{\mathbf{u}}_i = (\tilde{\mathbf{p}}_{ij},\tilde{\mathbf{q}}_{ij})$ is constructed $\mathcal{D}_{Q_i} = \{\tilde{\mathbf{s}}_{ij}\}_{j=1}^{N_{Q_i}}$ and used to fit a query density defined as 

\begin{ceqn}
\begin{equation}
\label{eq:query_density_approx}
Q_i(\mathbf{s}) \approx \sum_{k=1}^{K_{Q_i}} w_{k} \mathcal{N}(\mathbf{s} | \bm{\mu_s}^k,\bm{\Sigma_s}^k)
\end{equation}
\end{ceqn}

This is density is once again learnt as a GMM using the EM algorithm. 

In this fashion, the set of contact query models, one for each finger link and for a demonstrated grasp $g$, is denoted by $\mathcal{Q}_g=\{Q_i(\mathbf{s})\} \forall_i$.

\subsubsection{Grasp sampling}

The query density described above allows one to sample finger link poses over the given query point cloud surface. To construct a full hand posture, first a link pose is sampled using Eq. \ref{eq:query_density_approx}, i.e. $\tilde{\mathbf{s}} \sim Q_i(\mathbf{s})$. Finally, the full hand posture of a hand is completed using forward kinematics by sampling the remaining joint angles for the remaining links from the hand configuration model $C(\mathbf{h}_c)$. Many grasp candidates can be generated in this fashion.

\subsubsection{Grasp optimisation}

The generated grasp samples can be later optimised via a derivative-free method using the product of experts likelihood criterion as energy function, such as simulated annealing as done in \cite{kopicki2015}.

We denote a grasp solution as the tuple $\mathbf{h} = (\mathbf{h}_w, \mathbf{h}_c)$ containing respectively the hand wrist pose $\mathbf{h}_w \in SE(3)$ and hand joint configuration $\mathbf{h}_c \in \mathcal{R}^D$, such that if $fk(\cdot)$ is the forward kinematic function of the hand, then the hand links in workspace are given by $s_{1:N_L}$, where $N_L$ is the number of links of a robotic hand. Thus, we have

\begin{ceqn}
\begin{equation}
\label{eq:fk_function}
s_{1:N_L} = fk(\mathbf{h}), s_l = fk_l(\mathbf{h}).
\end{equation}
\end{ceqn}

The basic optimisation criterion is given by:

\begin{ceqn}
\begin{equation}
\label{eq:opt_criterion_std}
\mathcal{L}(\mathbf{h}) = C(\mathbf{h}) \prod_{Q_i \in \mathcal{Q}} Q_i(fk_i(\mathbf{h}))
\end{equation}
\end{ceqn}

\begin{ceqn}
\begin{equation}
\label{eq:opt_std}
\mathbf{\overset{*}{\mathbf{h}}} = \operatorname*{arg\,max}_{\mathbf{h}=(\mathbf{h}_w,\mathbf{h}_c)} = \mathcal{L}(\mathbf{h}).
\end{equation}
\end{ceqn}

In \cite{kopicki2015} an additional expert is defined $W(\mathbf{h})$ so as to penalise collisions in a soft manner, as it was found to prune the solution space to contain better solutions. Collisions are penalised exponentially by the degree of penetration through the object point cloud by any of the hand links. We have also employed a collision expert in the implementation of this work. 

The objective function with the additional collision expert is then given by

\begin{ceqn}
\begin{equation}
\label{eq:opt_criterion_std_col}
\mathcal{L}(\mathbf{h}) = W(\mathbf{h}) C(\mathbf{h}) \prod_{Q_i \in \mathcal{Q}} Q_i(fk_i(\mathbf{h})).
\end{equation}
\end{ceqn}

And again, the best grasp is found by optimising

\begin{ceqn}
\begin{equation}
\label{eq:opt_std_col}
\mathbf{\overset{*}{\mathbf{h}}} = \operatorname*{arg\,max}_{\mathbf{h}=(\mathbf{h}_w,\mathbf{h}_c)} \mathcal{L}(\mathbf{h})
\end{equation}
\end{ceqn}

\noindent where $W(\cdot)$ is the collision expert, $C(\cdot)$ is the hand configuration expert defined in Eq. \ref{eq:manipulator_model}, and $Q_i(\cdot)$ is the contact query expert for link $L_i$ defined by Eq. \ref{eq:query_density_approx}. For a robot hand and a given grasp demonstration $g$, there is a set of contact query experts $\mathcal{Q}_g$. This optimisation criterion, therefore, tries to maximise the product of experts \cite{Hinton99}, where each expert is a probability density. Each individual expert is responsible for assigning high likelihood to candidate grasps that satisfy just one of the constraints. 

In this work we will explore the mixture of experts approach even further and allow our optimisation criterion to incorporate multiple experts to represent arbitrary task constraints. This will give the potential to generate grasps specifically tailored for a given task. For this purpose let $\mathcal{E} = {E_k(\cdot)}_{k=1}^{N_k}$, such that each $E_i(\cdot)$ assigns a probability to a grasp solution $\mathbf{h}$ that is close to one if the grasp satisfies the constraint represented by $E_i(\cdot)$ and close to zero if the constraint is not satisfied. This set of constraints can be a set of constraints defined a priori by a human for a given task or learnt autonomously by the robot manipulator. We augment our optimisation criterion as follows:

\begin{ceqn}
\begin{equation}
\label{eq:opt_criterion_extended}
\mathcal{L}(\mathbf{h}) = W(\mathbf{h}) C(\mathbf{h}) \prod_{Q_i \in \mathcal{Q}} Q_i(fk_i(\mathbf{h})) \prod_{E_i \in \mathcal{E}} E_i(\mathbf{h})
\end{equation}
\end{ceqn}


Furthermore, we define a hard constraint as an expert that returns binary values of one or zero, accepting or rejecting instantly a grasp, thus implementing rejection sampling. Continuous experts whose image are on the continuous interval $[0,1]$ are referred to as soft constraints in this paper.

\begin{figure}[!t]
\centering
\includegraphics[width=0.75\columnwidth]{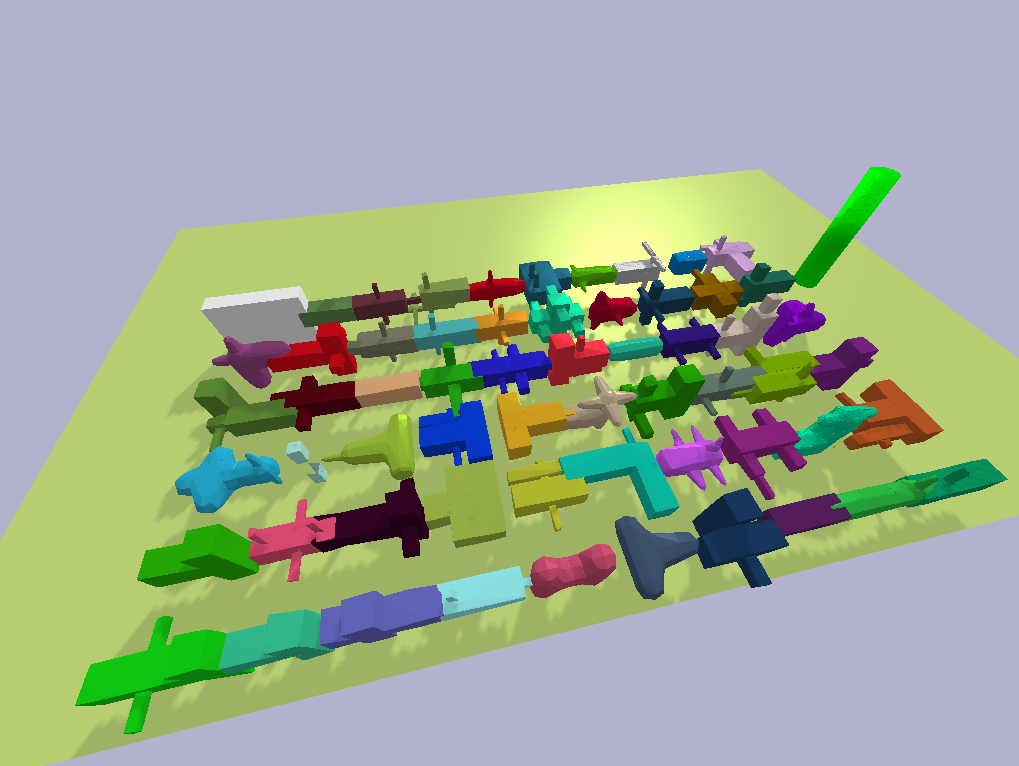}
\caption{\label{fig:objects} \small{Set of 60 test objects with varying shapes utilised for the simulated experiments.}}
\vspace{-12pt}
\end{figure} 

With this extension, we have then created a framework that allows a great variety of constraints to be incorporated into the grasp generation pipeline. For instance, one can think of experts that imbue dynamic constraints to grasps or specific kinematic constraints such as manipulability, thus paving the way towards task oriented grasp synthesis.

\begin{figure*}[!t]
\centering
\begin{minipage}{.5\textwidth}
  \centering
  \includegraphics[width=.5\linewidth]{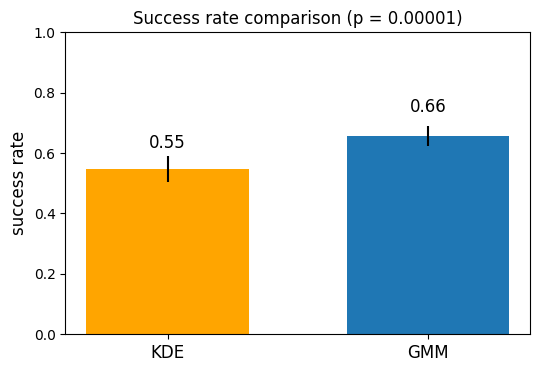}
  \captionof{figure}{\small{Condition A) No noise with optimisation.}}
  \label{fig:conditionANC}
  \vspace{4ex}
\end{minipage}
\begin{minipage}{.5\textwidth}
  \centering
  \includegraphics[width=.5\linewidth]{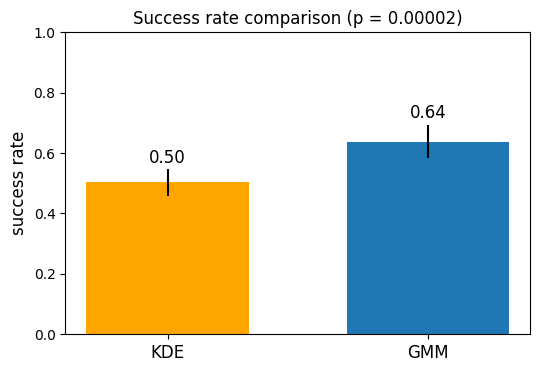}
  \captionof{figure}{\small{Condition B) With noise and optimisation.}}
  \label{fig:conditionBNC}
  \vspace{4ex}
\end{minipage}
\begin{minipage}{.5\textwidth}
  \centering
  \includegraphics[width=.5\linewidth]{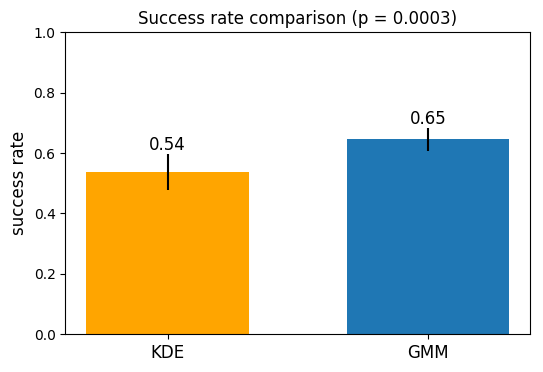}
  \captionof{figure}{\small{Condition C) No noise and without optimisation.}}
  \label{fig:conditionCNC}
    \vspace{-16pt}
\end{minipage}
\begin{minipage}{.5\textwidth}
  \centering
  \includegraphics[width=.5\linewidth]{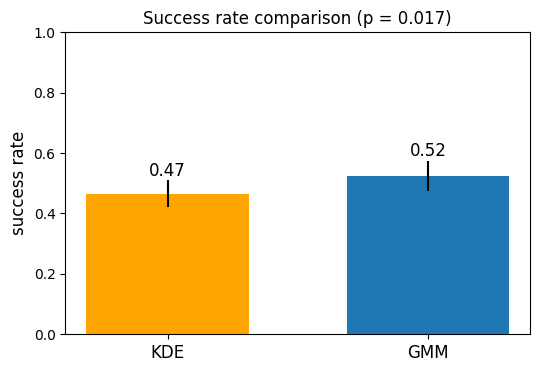}
  \captionof{figure}{\small{Condition D) With noise but no optimisation.}}
  \label{fig:conditionDNC}
    \vspace{-16pt}
\end{minipage}
\end{figure*}

Thus far in this work, the following constraint experts have been defined: i) hard kinematic constraint and ii) soft principal axis alignment constraint. The former expert returns zero if a given grasp solution $\mathbf{h}$ is not kinematically feasible by the robot platform, and one otherwise. It implements a hard constraint over the grasp generation process, rejecting grasp samples that are not kinematically feasible. The latter expert implements a soft constraint that returns a probability close to one if the wrist frame of the generated grasp $\mathbf{h}$ is aligned with a given axis provided by the user. This is useful to encourage grasps that always pick an object from the top, for instance.

\section{Experiments}

\begin{figure}[h]
\vspace{4pt}
\centering
\includegraphics[width=0.90\columnwidth]{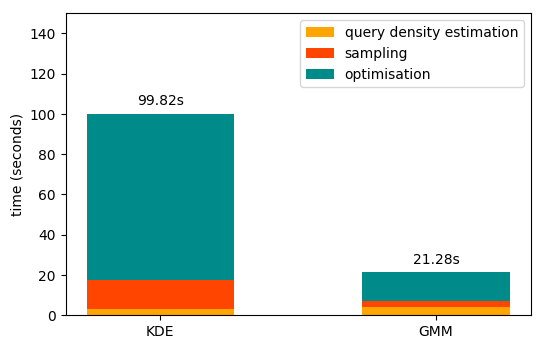}
\caption{\small{Corresponding total runtime for KDE and GMM-based methods. The total time includes the time for query density estimation, sampling of 200 grasps and optimisation of top 10 grasps using 100 iterations of simulated annealing. The results the GMM-based method is nearly 5 times faster. The GMM-based method requires fewer kernels to approximate the query density and therefore is much faster for likelihood evaluation during optimisation. For the same reason this gain in time performance is also noted during sampling and evaluation.}}
\label{fig:runtime} 
\vspace{-12pt}
\end{figure}

We perform two separate experiments. The first experiment compares the success rate between the KDE and GMM-based success rate in simulation (we used Bullet physics \cite{Coumans2019}) in four conditions: A) without depth sensing noise and with optimisation, B) with simulated depth sensing noise and with optimisation, C) without noise and without optimisation and D) with noise and without optimisation. The second experiment shows the runtime performance for each respective method when generating 200 hundred grasps for an object, results are shown in Fig. \ref{fig:runtime}.

For all experiments and grasp synthesis methods, the optimisation utilised was the derivative-free simulated annealing method with 100 iterations and maximum initial temperature of 1.0 which decays linearly over the number of iterations.

We adopted the same depth sensor noise model as in \cite{Johns2016}. The noise model combines additive and lateral shift Gaussian noise applied to every pixel in a simulated depth image:

\begin{equation}
\label{eq:noise_model}
\tilde{z}(u,v) = z(u + \mathcal{N}(0, \sigma_p^2), v + \mathcal{N}(0, \sigma_p^2)) + \mathcal{N}(0, \sigma_d^2),
\end{equation}

\noindent where $\sigma_p$ is the standard deviation for the noisy lateral shift of pixels, which has been set to 1 for these experiments. And $\sigma_d$ is the standard deviation for depth measurements, which was set to 1mm.

\subsection{Performance comparison}

For the comparison, we use a parallel gripper (WSG 50 Shunk Gripper) and a single demonstrated grasp on a box, from which object the contact models are learnt for both KDE and GMM-based methods. The KDE-based method has utilised one kernel per data point to approximate the densities. The GMM-based contact model has approximated Gaussian mixture densities with $K=2$, given the bi-modality of the demonstration as depicted in Fig. \ref{fig:cd_feat} 
and with the number of kernels $K_{Q}=5$ for GMM-based query density approximation from Eq. \ref{eq:query_density_approx}. 

A total of 60 objects were used for testing the performance of each respective method. Objects are varied, some objects have box-like features, others are curved, cylindrical, convex, non-convex, and of combinations of sizes and shapes as shown in Fig. \ref{fig:objects}.
Objects are always placed in the same pose for both approaches, and simulation parameters are kept fixed. For grasp generation, simulated point clouds are acquired from 4 fixed camera poses around the object, thus obtaining a full 3D point cloud of the object. The acquired point clouds are stitched together and post-processed identically for both approaches, including principal curvature feature computation with search radius set to $1cm$.

For each respective approach (KDE and GMM-based), a total number of 200 grasps are generated per object. Sampled grasp solutions are arranged in decreasing order of value as given by Eq. \ref{eq:opt_criterion_std}, in which the highest value represents the best grasp in the sequence. For each approach, the best grasp is executed on each object. Before every execution attempt the object is restarted to the same initial pose.
This process is repeated 10 times over the set of 60 object and results are averaged over these 10 trials. We measure success rate as the proportion of successful grasps over the total number of objects.

Finally, a two-tailed paired t-test is performed to assert the significance of the results between KDE and GMM-based methods over the 10 repeated trials.

\subsection{Preliminary tests on robot platform}

We have preliminarily deployed the proposed approach on a real robot platform as shown by Fig. \ref{fig:cylinder_grasp}. For the robot platform, we made use task-specific kinematic constraints, such as kinematic feasibility and axis alignment that can be straightforwardly added in our implementation using Eq. \ref{eq:opt_criterion_extended} to prune the grasp synthesis process. A summary video on the Boris platform is available on \url{https://youtu.be/9RqZaTAH4Fs}.

\subsection{Results}

The results in all four different conditions indicate that the GMM-based method shows superior performance over the KDE-based method ($p < 0.05$) in this setup. We see that in the absence of noise, the final optimisation step has little effect in grasp performance as shown by Fig. \ref{fig:conditionANC} and \ref{fig:conditionCNC}. In contrast, optimisation seems to become crucial under noisy conditions as depicted by Fig. \ref{fig:conditionBNC} and \ref{fig:conditionDNC}, which seems natural. Under the influence of noise, local optimisation with simulated annealing is expected to help to compensate from imperfect sensing, hence still converging to a local or global minimum under the product of experts objective function in Eq. \ref{eq:opt_criterion_std}. 

Perhaps the biggest impact is shown in terms of the total time required for grasp synthesis. The stacked bar chart in Fig. \ref{fig:runtime} shows the total runtime for generating 200 grasps for a given object, highlighting the individual contributions from different steps involved in the grasp synthesis process: i) query density estimation, ii) sampling and ranking and finally iii) optimising top 10 grasps (out of the 200 samples). Due to the fact that the GMM-based method requires fewer kernels to approximate the query density in Eq. \ref{eq:query_density_approx}, it directly affects the total runtime which is largely dominated by the number of query likelihood evaluations (Eq. \ref{eq:opt_criterion_std}).

\subsection{Conclusion \& Future Work}

We proposed an alternative formulation for learning generative models for grasp synthesis using parametric mixtures, such as GMMs. In our experimental setup, the GMM-based method has indicated better performance than related non-parametric approach using KDEs \cite{kopicki2015}. In addition to that, we extended the product of experts criterion, allowing us to accommodate task-specific constraints as needed for different deployment contexts. We gave a brief instance of that by including kinematic constraints directly in the evaluation step of the grasp synthesis pipeline on the Boris platform. Finally, we showed the GMM-based approach also offers faster execution times. This is a promising prospect that may allow us to employ the proposed method to real-time grasp synthesis as future work. Subsequent work in this direction shall be conducted on a real robot platform.


\addtolength{\textheight}{-12cm}   

\bibliographystyle{plain}
\bibliography{grasp_synthesis,general_ml,simulation}

\end{document}